\documentclass[10pt,twocolumn,letterpaper]{article}

\usepackage{cvpr}
\usepackage{times}
\usepackage{epsfig}
\usepackage{graphicx}
\usepackage{amsmath}
\usepackage{amssymb}
\usepackage{subfigure}
\usepackage{multirow}

\usepackage[breaklinks=true,bookmarks=false]{hyperref}

\cvprfinalcopy 

\def\cvprPaperID{1255} 

\newtheorem{condition}{Condition}
\begin{document}
\setlength{\abovedisplayskip}{3pt}
\setlength{\belowdisplayskip}{3pt}
\setlength{\abovedisplayshortskip}{0pt}
\setlength{\belowdisplayshortskip}{0pt}
\title{IGCV$2$: Interleaved Structured Sparse Convolutional Neural Networks}

\author{Guotian Xie$^{1,2,}$\thanks{This work was done when Guotian Xie was an intern at Microsoft Research, Beijing, P.R. China}~~~ Jingdong Wang$^3$
~~ Ting Zhang$^3$~~ Jianhuang Lai$^{1,2}$~~ Richang Hong$^4$~~ Guo-Jun Qi$^5$\\
$^1$Sun Yat-Sen University~~ $^2$Guangdong Key Laboratory of Information Security Technology\\
$^3$Microsoft Research~~
$^4$Hefei University of Technology~~
$^5$University of Central Florida\\
{\tt\small xieguotian1990@gmail.com, \{jingdw,tinzhan\}@microsoft.com}\\
{\tt\small stsljh@mail.sysu.edu.cn, hongrc.hfut@gmail.com, guojun.qi@ucf.edu}
}

\maketitle

\begin{abstract}
In this paper, we study the problem
of designing efficient convolutional neural network architectures
with the interest in eliminating the redundancy
in convolution kernels.
In addition to
structured sparse kernels,
low-rank kernels
and the product of low-rank kernels,
the product of structured sparse kernels,
which is a framework for
interpreting the recently-developed
interleaved group convolutions (IGC)
and its variants (e.g., Xception),
has been attracting increasing interests.

Motivated by the observation
that the convolutions contained in a group convolution in IGC
can be further decomposed in the same manner,
we present a modularized building block,
{IGCV$2$:} interleaved structured sparse convolutions.
It generalizes interleaved group convolutions,
which is composed of two structured sparse kernels,
to the product of more structured sparse kernels,
further eliminating the redundancy.
We present the complementary condition
and the balance condition
to guide the design of structured sparse kernels,
obtaining a balance among three aspects: model size, computation complexity
and classification accuracy.
Experimental results
demonstrate the advantage on
the balance among these three aspects
compared to interleaved group convolutions and Xception,
and competitive performance
compared to other state-of-the-art architecture design methods.
\end{abstract}

\section{Introduction}

Deep convolutional neural networks
with small model size, low computation cost,
but still high accuracy
become an urgent request,
especially in mobile devices.
The efforts include
(i) network compression:
compress the pretrained model
by decomposing the convolutional kernel matrix
or removing connections or channels
to eliminate redundancy,
and (ii) architecture design:
design small kernels,
sparse kernels
or use the product of less-redundant kernels
to approach single kernel
and train the networks from scratch.

Our study lies in architecture design
using
the product of less-redundant kernels
for composing a kernel.
There are two main lines:
multiply low-rank kernels (matrices)
to approximate a high-rank kernel,
e.g., bottleneck modules~\cite{HeZRS16},
and multiply sparse matrices,
which has attracted research efforts recently~\cite{ZhangQXW17, IoannouRCC16, Chollet16a}
and is the focus of our work.

We point out that the recently-developed algorithms,
such as
interleaved group convolution~\cite{ZhangQXW17}, deep roots~\cite{IoannouRCC16}, and Xception~\cite{Chollet16a},
compose a dense kernel
using the product of two structured-sparse kernels.
We observe that
one of the two kernels can be further approximated.
For example,
the $1 \times 1$ kernel in Xception and deep roots
can be approximated
by the product of two block-diagonal sparse matrices.
The suggested secondary group convolution
in interleaved group convolutions
contains two branches
and each branch is a $1\times 1$ convolution,
which similarly can be further approximated.
This is able to further reduce the redundancy.

\begin{figure*}
	\centering
		\includegraphics[width=0.85\textwidth]{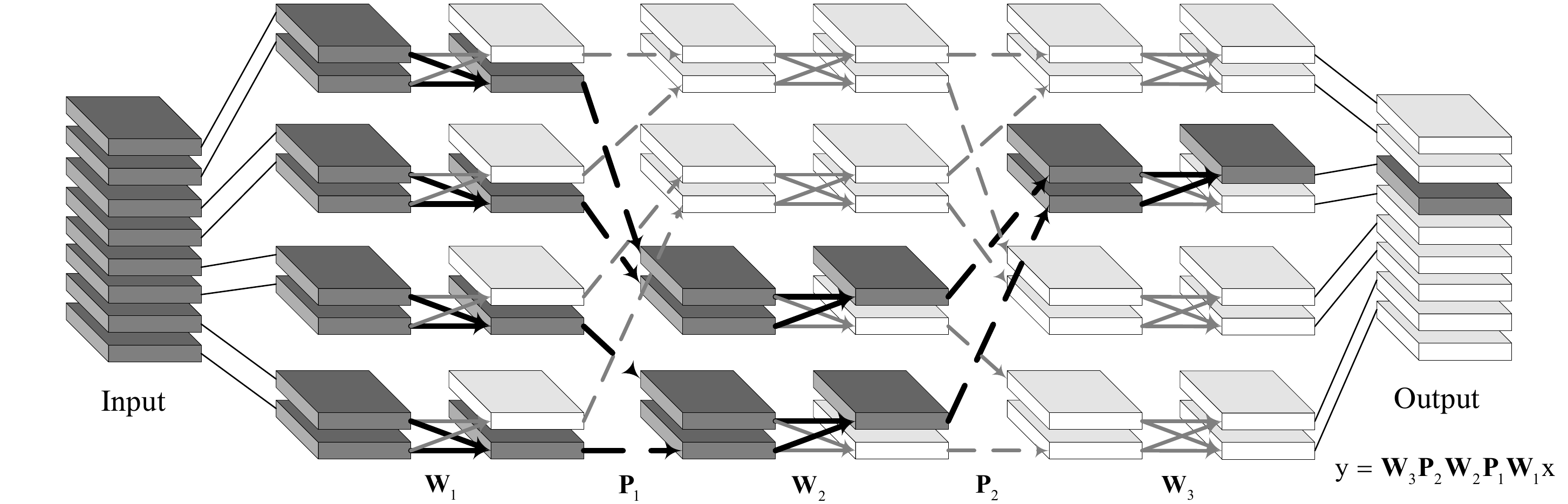}
		\caption{
			{IGCV2: the Interleaved Structured Sparse Convolution.}
			$\mathbf{W}_1$, $\mathbf{W}_2$, $\mathbf{W}_3$ (denoted as solid arrows) are sparse block matrices
			corresponding to group convolutions. $\mathbf{P}_1$ and $\mathbf{P}_2$ (denoted as dashed arrows) are
			permutation matrices. The resulting composed kernel
			$\mathbf{W}_3\mathbf{P}_2\mathbf{W}_2\mathbf{P}_1\mathbf{W}_1$ is ensured to
			satisfy the {\emph {complementary condition}} which guarantees
			that for each output channel, there exists one
			and only one path connecting the output channel to each input channel.
			The bold line connecting gray feature maps shows such a path. 			
		}
		\label{fig:ISSC}
		\vspace{-0.5cm}
\end{figure*}

Motivated by this,
we design a building block, {IGCV$2$:}
Interleaved Structured Sparse Convolution, as shown in Figure~\ref{fig:ISSC},
which consists of successive group convolutions.
This block is mathematically
formulated as
multiplying structured-sparse kernels,
each of which corresponds to a group convolution.
We introduce the complementary condition
and the balance condition,
so that the resulting convolution kernel is dense
and there is a good balance
among three aspects: model size, computation complexity
and classification performance.
Experimental results
demonstrate the advantage of
the balance among these three aspects
compared to interleaved group convolutions and Xception,
and competitive performance
compared to other state-of-the-art architecture design methods.

\section{Related Work}
Most existing technologies design
efficient and effective
convolutional kernels
using various forms with redundancy eliminated,
by learning from scratch or
approximating pretrained models.
We roughly divide them into
low-precision kernels,
sparse kernels,
low-rank kernels,
product of low-rank kernels,
product of structured sparse kernels.

\vspace{0.1cm}
\noindent\textbf{Low-precision kernels.}
There exist redundancies
in the weights in convolutional kernels
represented by float numbers.
The technologies eliminating such redundancies
include quantization~\cite{zhou2017incremental,HanMD15},
binarization~\cite{courbariaux2016binarized,rastegari2016xnor},
and
trinarization~\cite{li2016ternary,zhou2016dorefa,zhu2016trained}.
Weight-shared kernels
in which some weights are equal to the same value,
are in some sense
low-precision kernels.

\vspace{0.1cm}
\noindent\textbf{Sparse kernels.}
Sparse kernels,
or namely sparse connections,
mean that
some weights are nearly zero.
The efforts along this path
mainly lie in how to perform optimization,
and the technologies include
non-structure sparsity regularization~\cite{liu2015sparse,park2016faster},
and structure sparsity regularization~\cite{WenWWCL16,mao2017exploring}.
The scheme of structure sparsity regularization is more friendly for hardware acceleration
and storage.
Recently, group convolutions,
adopted in~\cite{XieGDTH16, ZhaoWLTZ16}
are essentially
structured-sparse kernels.
Different from sparsity regularization,
the sparsity pattern of group convolution is manually pre-defined.

\vspace{0.1cm}
\noindent\textbf{Low-rank kernels.}
Small filters, e.g., $3 \times 3$ kernels
replacing $5 \times 5$ kernels,
reduce the ranks in the spatial domain.
Channel pruning~\cite{LiuLSHYZ17} and
filter pruning~\cite{wen2017coordinating,LiKDSG16,LuoWL17}
compute low-rank kernels
in the output channel domain
and the input channel domain, respectively\footnote{Small filters, channel and filter pruning
	in some sense can also be interpreted as
	sparse kernels:
	some columns or rows are removed.}.

\vspace{0.1cm}
\noindent\textbf{Composition from low-rank kernels.}
Using a pair of
$1 \times 3$ and $3 \times 1$ kernels
to approximate a $3\times 3$ kernel~\cite{IoannouRSCC15, JaderbergVZ14, MamaletG12}
is an example
of using the product of two small (low-rank) filters.
Tensor decomposition uses
the product of low-rank/small tensors (matrices) to approximate
the kernel in the tensor form
along
the spatial domain~\cite{DentonZBLF14, JaderbergVZ14},
or the input and output channel domains~\cite{DentonZBLF14, KimPYCYS15, JaderbergVZ14}.
The bottleneck structure~\cite{HeZRS16},
if the intermediate ReLUs are removed,
can be viewed as the low-rank approximation
along the output channel domain.

\vspace{0.1cm}
\noindent\textbf{Composition from sparse kernels.}
Interleaved group convolution~\cite{ZhangQXW17} consists of
two group convolutions,
each of which corresponds to a structured-sparse kernel
(the sizes are the same to that of the kernel to be approximated
for the $1\times 1$ convolutions).
Satisfying the complementary property~\cite{ZhangQXW17}
leads to that
the resulting composite kernel is dense.
Xception~\cite{Chollet16a} can be viewed
as an extreme case of interleaved group convolutions:
one group convolution is degraded to a regular convolution
and the other one is a channel-wise convolution,
an extreme group convolution.
Deep roots~\cite{IoannouRCC16}
instead uses the product of a structured-sparse kernel and a dense kernel.
Our approach belongs to this category
and shows a better balance among model size, computation complexity
and classification accuracy.

\section{Our Approach}
The operation in a convolution layer
in convolutional neural networks
relies on
a matrix-vector multiplication operation at each location:
\begin{align}
\mathbf{y} = \mathbf{W} \mathbf{x}.
\end{align}
Here the input $\mathbf{x}$,
corresponding to a patch around the location
in the input channels,
is a $SC_i$-dimensional vector,
with $S$ being the kernel size (e.g., $S= 3 \times 3$),
$C_i$ being the number of input channels.
The output $\mathbf{y}$
is a $C_o$-dimensional vector,
with $C_o$ being the number of output channels.
$\mathbf{W}$ is formed
from $C_o$ convolutional kernels
and each row corresponds to a convolutional kernel.
For presentation clarity,
we assume $C_i=C_o = C$,
but all the formulations can be generalized
to $C_i \neq C_o$.

\subsection{A Review of IGC, Xception and Deep Roots}
We show that recent architecture design algorithms,
Xception~\cite{Chollet16a},
deep roots~\cite{IoannouRCC16},
and interleaved group convolutions (IGC)~\cite{ZhangQXW17},
compose a dense convolution matrix $\mathbf{W}$
by multiplying possibly sparse matrices:
\begin{align}
\mathbf{y} = \mathbf{P}^2\mathbf{W}^2 \mathbf{P}^1\mathbf{W}^1 \mathbf{x},
\label{eqn:twosparsematrixproduct}
\end{align}
where $\mathbf{W}^1$ and $\mathbf{W}^2$
are both,
or at least one matrix is block-wise sparse,
$\mathbf{P}^i$ is a permutation matrix
that is used to reorder the channels,
and $\mathbf{W} = \mathbf{P}^2\mathbf{W}^2 \mathbf{P}^1\mathbf{W}^1$ is a dense matrix.

\vspace{.1cm}
\noindent\textbf{Interleaved group convolutions.}
The interleaved group convolution block
consists of primary and secondary group convolutions.
The corresponding kernel matrices $\mathbf{W}^1$ and $\mathbf{W}^2$ are block-wise sparse,
\begin{align}
\mathbf{W}^i =  \begin{bmatrix}
\mathbf{W}_{1}^i & \boldsymbol{0} &  \boldsymbol{0} &  \boldsymbol{0} \\[0.3em]
\boldsymbol{0} & \mathbf{W}_{2}^i  & \boldsymbol{0} & \boldsymbol{0} \\[0.3em]
\vdots & \vdots & \ddots  & \vdots \\[0.3em]
\boldsymbol{0} & \boldsymbol{0} & \boldsymbol{0} & \mathbf{W}_{G_i}^i
\end{bmatrix},
\label{eqn:groupconvolution}
\end{align}
where $\mathbf{W}_{g}^i$ ($i=1$ or $2$) is
the kernel matrix
over the corresponding channels
in the $g$th branch,
$G_i$ is the number of branches
in the $i$th group convolution.
In the case suggested in~\cite{ZhangQXW17},
the primary group convolution is a
group $3\times 3$
convolution,
$G_1 = \frac{C}{2}$,
and $\mathbf{W}^1_g$ is a matrix of size $2 \times (2S)$.
The secondary group convolution is a group $1\times 1$ convolution,
$G_2 = 2$,
$\mathbf{W}^2_1$ and $\mathbf{W}^2_2$
are both dense matrices of size $\frac{C}{2} \times \frac{C}{2}$.

\vspace{.1cm}
\noindent\textbf{Xception.}
The Xception block consists of
a $1\times 1$ convolution layer
followed by a channel-wise convolution layer.
It is pointed out that the order of the two layers does not make effects.
For convenience, we below discuss the form
with the $1 \times 1$ convolution put as the second operation.
$\mathbf{W}^2$ is a dense matrix of size $C \times C$.
$\mathbf{W}^1$ is a sparse block matrix of size $C \times (SC)$,
a degraded form of the matrix shown in Equation~\ref{eqn:groupconvolution}:
there are $C$ blocks
and $\mathbf{W}^1_{g}$ is degraded
to a row vector of size $S$.

\vspace{.1cm}
\noindent\textbf{Deep roots.}
In deep roots,
$\mathbf{W}^2$ is a dense matrix of size $C \times C$,
i.e., corresponding to a $1 \times 1$ convolution
while $\mathbf{W}^1$ is a sparse block matrix
as shown in Equation~\ref{eqn:groupconvolution},
corresponding to a group convolution.

\vspace{.1cm}
\noindent\textbf{Complexity.}
The computation complexity
of Equation~\ref{eqn:twosparsematrixproduct}
is $O(|\mathbf{W}^1|_0 + |\mathbf{W}^2|_0)$
(with the complexity in permutation
is ignored),
where $|\mathbf{W}^i|_0$
is the number of non-zero entries.
The sparse block matrix,
as given in Equation~\ref{eqn:groupconvolution},
are storage friendly,
and
the storage/memory cost is also $O(|\mathbf{W}^1|_0 + |\mathbf{W}^2|_0)$.

\subsection{Interleaved Structured Sparse Convolutions}
\label{sec:ISSC}
Our approach is motivated by
the observations:
(i) the block $\mathbf{W}_g^i$ in Equation~\ref{eqn:groupconvolution}
and the $1\times 1$ convolution in Xception
are dense and can be composed
by multiplying sparse matrices,
thus further eliminating the redundancy
and saving the storage and time cost;
and (ii)
such a process can be repeated more times.

The proposed Interleaved Structured Sparse Convolution ({IGCV$2$})
is mathematically formulated as follows,
\begin{align}
\mathbf{y} =~& \mathbf{P}_L\mathbf{W}_L \mathbf{P}_{L-1}\mathbf{W}_{L-1} \dots \mathbf{P}_1\mathbf{W}_1 \mathbf{x} \\
=~& (\prod\nolimits_{l=L}^1 \mathbf{P}_l \mathbf{W}_l) \mathbf{x}.
\label{eqn:moresparsematrixproduct}
\end{align}
Here, $\mathbf{P}_l \mathbf{W}_l$
is a sparse matrix.
$\mathbf{P}_l$ is a permutation matrix,
and the role is to reorder the channels
so that $\mathbf{W}_l$
is a sparse block matrix,
as given in Equation~\ref{eqn:groupconvolution}
and corresponds to the $l$th group convolution,
where the numbers of channels in all the branches
are in our work set to be the same,
equal to $K_l$,
for easy design.

\vspace{.1cm}
\noindent\textbf{Construct a dense composed kernel matrix.}
We introduce the following~\emph{complementary condition},
which is generalized from interleaved group convolutions~\cite{ZhangQXW17},
as a rule for constructing the $L$ group convolutions
such that the resulting composed convolution kernel matrix
is~\emph{dense}.

\begin{condition}[Complementary condition]
	$\forall m$,
	$(\mathbf{W}_L\prod_{l=L-1}^m \mathbf{P}_l \mathbf{W}_l)$
	corresponds to a group convolution
	and
	$(\mathbf{W}_{m-1}\prod_{l=m-2}^1 \mathbf{P}_l \mathbf{W}_l)$
	also corresponds to a group convolution.
	The two group convolutions are thought complementary
	if
	the channels lying in the same branch
	in one group convolution
	lie in different branches
    and come from all the branches in the other group convolution.

\end{condition}

Here is the sketch showing
that an interleaved structured sparse convolution block satisfying the complementary condition
is dense.
The proof is based on two points:
(i) for a group convolution,
we have that any channel output from a branch is connected to
the channels input to this branch
and any channel input to a branch
is connected to the channels output from this branch;
(ii) for two complementary group convolutions,
the channels output from any branch of the second group convolution
are connected to
the channels input to the corresponding branch,
which are from all the branches of the first group convolution.
As a result,
the channels output from an {IGCV$2$}
is connected to all the channels input to the {IGCV$2$},
i.e., the {IGCV$2$} kernel is~\emph{dense}.

Let us look at the relation
between the number of channels, $C$,
and the number of channels in the branches
of $L$ group convolutions,
$\{K_1, K_2, \dots, K_L\}$.
We analyze the relation according to~Equation~\ref{eqn:moresparsematrixproduct}:
(i)
An input channel is connected to $K_1$ intermediate channels
output by the first group convolution.
(ii) Let $C_{l-1}$
be the number of intermediate channels
output by the $(l-1)$th group convolution,
to which an input channel is connected.
The \emph{complementary condition} indicates that
through the $l$th group convolution
an input channel is connected
to \emph{exactly} $K_lC_{l-1}$ intermediate channels
output by the $l$th group convolution.
(iii) Finally,
an input channel is connected
to exactly $C_L = \prod_{l=1}^L K_l$ channels
output from the $L$ group convolutions.
Since the composed kernel is dense,
we have
\begin{align}
\prod\nolimits_{l=1}^L K_l = C.
\label{eqn:multiplicationproperty}
\end{align}

Because of the complementary property,
there is no waste connection:
there is only one path between each input channel
and each output channel.
Besides, the complementary condition
is a sufficient condition yielding a dense composed kernel matrix,
and not a necessary condition.

\vspace{.1cm}
\noindent\textbf{When the amount of parameters is the smallest?}
We further analyze when the number of parameters with $L$ group convolutions,
as given in Equation~\ref{eqn:moresparsematrixproduct},
satisfying the complementary condition,
is the smallest.

We have that
the number of parameters in the $l$th group convolution
is $CK_l$
for the $1\times 1$ convolutions,
and $CSK_l$
for the spatial (e.g., $S = 3 \times 3$) convolution.
It is easily shown that
for consuming fewer parameters
there is only one group spatial convolution
and all others are $1\times 1$.
The spatial convolution lies in any group convolution,
and without affecting the analysis,
we assume it lies in
the first group convolution\footnote{The formulation is not the same to~Equation~\ref{eqn:moresparsematrixproduct}
	if the group $3 \times 3$ convolution is not the first,
	but essentially they are the same.}.
Thus, the number of total parameters $Q$,
smaller number of parameters in permutation matrices ignored,
is:
\begin{align}
Q = C\sum\nolimits_{l=2}^{L} K_l + CSK_1.
\end{align}
According to Jensen's inequality,
we have
\begin{align}
Q =~& C\sum\nolimits_{l=2}^{L} K_l + CSK_1 \\
\geqslant ~& C L (SK_1\prod\nolimits_{l=2}^L K_l)^{\frac{1}{L}} \\
=~& CL(SC)^{\frac{1}{L}}.
\end{align}
Here, the equality from the second line to the third line
holds
because of Equation~\ref{eqn:multiplicationproperty}.
The equality in the second line holds,
i.e., $Q = CL(SC)^{\frac{1}{L}}$,
when the following~\emph{balance condition} is satisfied\footnote{This can be regarded as an extension
	of the analysis in~\cite{ZhangQXW17}.},
\begin{align}
SK_1 = K_2 = \dots = K_L (=(SC)^{\frac{1}{L}}).
\label{eqn:balancecondition}
\end{align}

Furthermore,
let us see
the choice of $L$,
yielding the smallest amount of parameters ($Q = CL(SC)^{\frac{1}{L}}$),
guaranteeing a dense composed kernel.
We present a rough analysis
by considering the derivative
of $Q$ with respect to $L$:
\begin{align}
\frac{d \log Q}{dL} =~&
\frac{d }{d L} (\log C + \log L + \frac{1}{L}\log (SC)) \\
= ~& \frac{1}{L} - \frac{1}{L^2} \log (SC).
\end{align}
When the derivative is zero,
$\frac{1}{L} - \frac{1}{L^2} \log (SC) = 0$,
we have that $Q$ is the minimum if
\begin{align}
L = \log (SC).
\label{eqn:optimalLforminimalC}
\end{align}


\vspace{.1cm}
\noindent\textbf{Examples.}
We take an example:
separate the convolution along the spatial domain
and the channel domain,
to construct the {IGCV$2$} block.
The first group convolution is an extreme group convolution,
a channel-wise $3 \times 3$ convolution,
followed by several group $1 \times 1$ convolutions.
This can be regarded
as decomposing the $1\times 1$ convolution in Xception
into group $1\times 1$ convolutions.
In this case,
the balance condition becomes
$K_2=K_3= \dots = K_L = C^{\frac{1}{L-1}}$,
for which the amount of parameters is the smallest.
As we empirically validate in Section~\ref{sec:comparisonwithxception},
under the same number of parameters,
an {IGCV$2$} block satisfying such a balance condition
leads to the maximum width
and
consistently superior performance:
the best or nearly best.
This consistency observation is different from~\cite{ZhangQXW17}
and might stem from
that the balance condition is only used to $1\times 1$ group convolutions
and that there is no coupling with spatial convolutions.

We also study the construction
from interleaved group convolutions:
each submatrix $\mathbf{W}^2_g$ in Equation~\ref{eqn:groupconvolution}
in the secondary group convolution
corresponds to a (dense) $1 \times 1$ convolution
over a subset of channels,
and thus can be further decomposed into group convolutions.
The first group convolution is still a group $3 \times 3$ convolution
(other than channel-wise).
Consequently, the balance condition given in Equation~\ref{eqn:balancecondition}
is deduced from the coupling of convolutions
over the spatial and channel domains,
which does not lead to the consistency
between the width increase and the performance gain
and makes the analysis uneasy.
This is empirically validated in our experiments in Section~\ref{sec:comparisonwithIGC}.
Thus, we suggest
to separate the convolution along
the spatial and channel domains
and design an {IGCV$2$}
over the channel domain.

\begin{figure*}
	\centering
	{\small (a)}\includegraphics[width=0.3\textwidth]{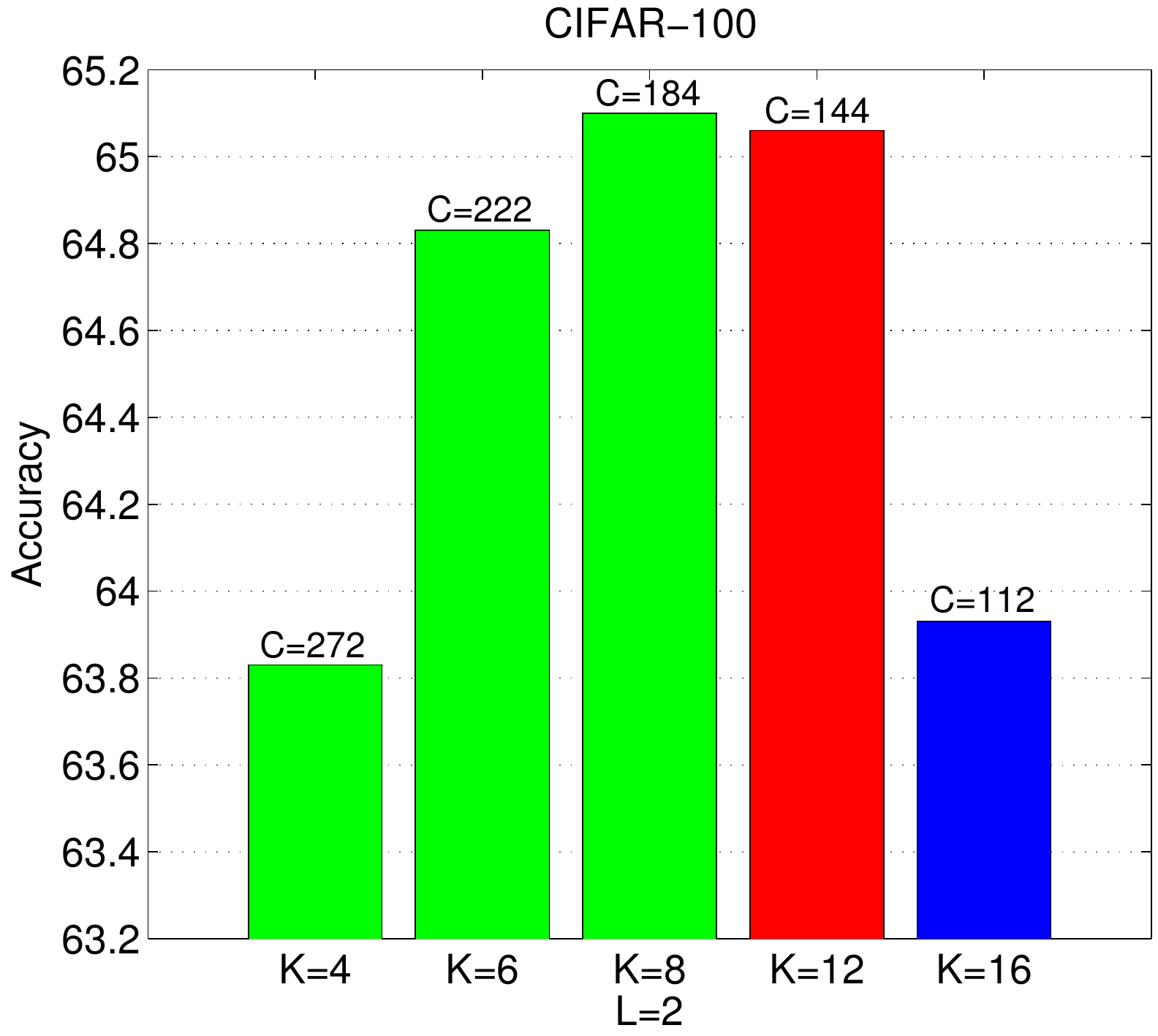}~~
	{\small (b)}\includegraphics[width=0.3\textwidth]{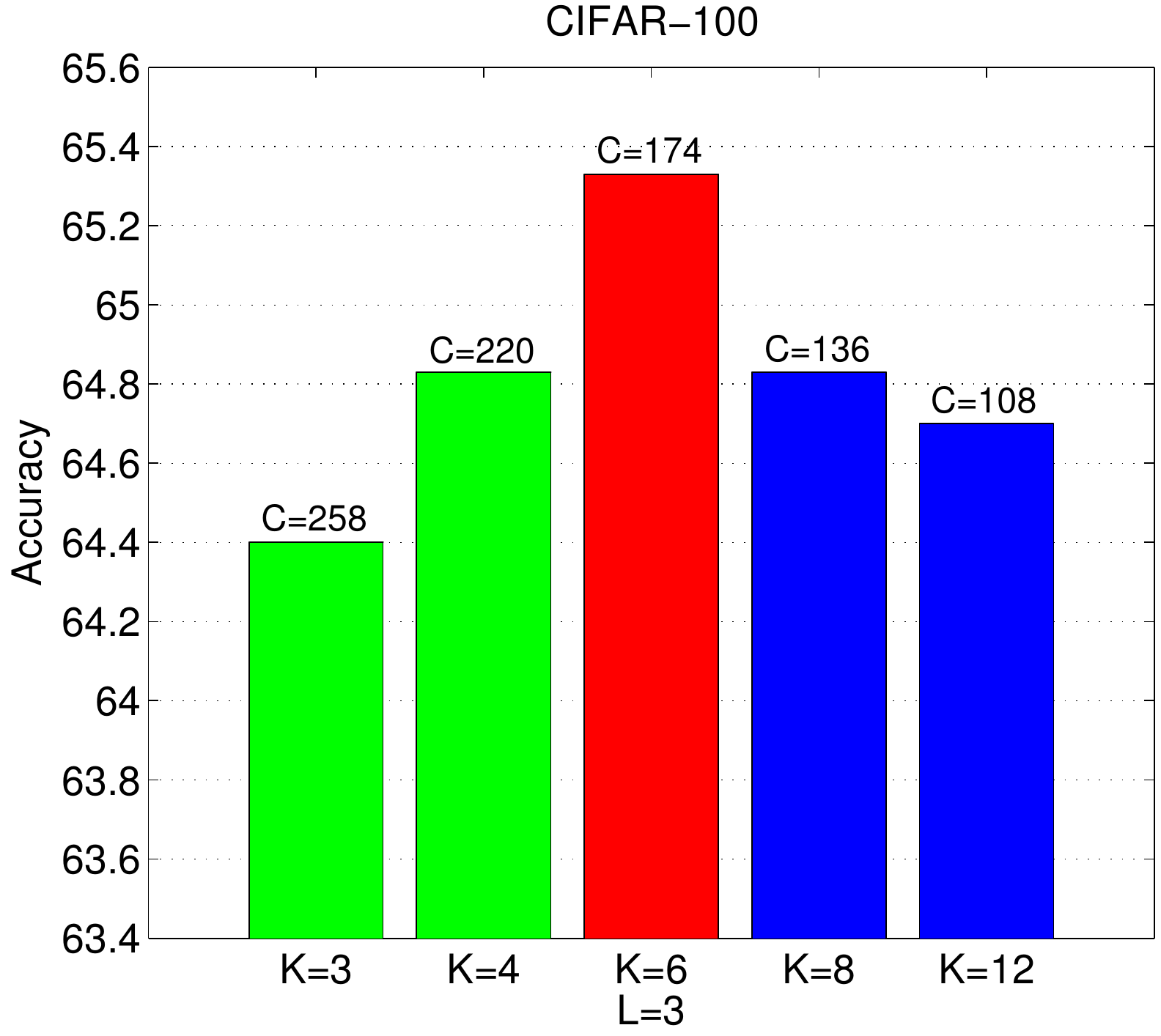}~~
	{\small (c)}\includegraphics[width=0.3\textwidth]{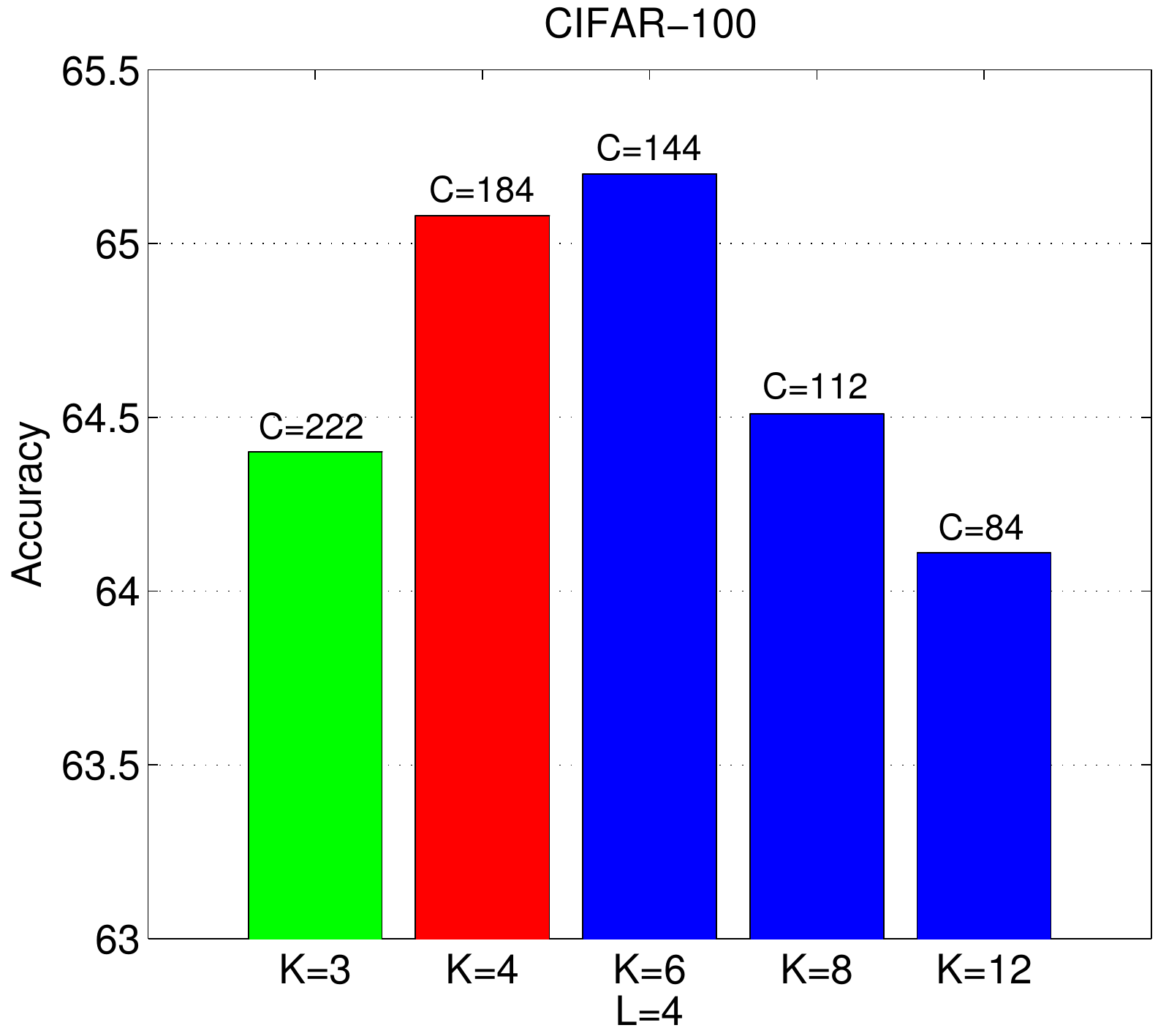}
	\caption{ Illustrating how the complementary condition affects the performance on CIFAR-$100$ in our approach.
		$K$ denotes the number of channels in each branch and $C$ denotes the width of the network.
		 With $L$ fixed, the composed kernel is denser with a larger $K$.
		The red bar corresponds to the case in which the complementary condition
		is the most satisfied.
		The best performances corresponding to the red bar or
		the bars immediately near to the red bar
		show that the complementary condition is reasonable
		for {IGCV$2$} design.}
	\label{fig:effectofk}
	\vspace{-0.4cm}
\end{figure*}

\subsection{Discussions}
\noindent\textbf{Non-structured sparse kernels.}
There is a possible extension:
remove the structured sparsity requirement,
i.e., replace the group convolution
by a non-structured sparse kernel,
and introduce the dense constraint
(the composed kernel is dense)
and the sparsity constraint.
This potentially results in better performance,
but leads to two drawbacks:
the optimization is difficult
and non-structured sparse matrices
are not storage-friendly.

\vspace{.1cm}
\noindent\textbf{Complementary condition.}
The complementary condition is a sufficient
condition guaranteeing the resulting composed kernel is dense.
It should be noted that
it is not a necessary condition.
Moreover,
it is also not necessary that the composed kernel is dense,
and further sparsifying the connections,
which remains as a future work, might be beneficial.
The complementary condition
is an effective guide to design the group convolutions.

\vspace{.1cm}
\noindent\textbf{Sparse matrix multiplication and low-rank matrix multiplication.}
Low-rank matrix (tensor) multiplication
or decomposition
has been widely studied in matrix analysis~\cite{LeeKLS13, LeeKLSB16}
and applied to network compression
and network architecture design.
In comparison,
sparse matrix (tensor) multiplication or decomposition
is rarely studied in matrix analysis.
The future works include applying sparse matrix decomposition
to compress convolutional networks,
combining low-rank and sparse matrices together: low-rank sparse matrix multiplication or decomposition,
and so on.

\section{Experiment}

\subsection{Datasets and Training Settings}
\noindent\textbf{CIFAR.}
The CIFAR datasets~\cite{Alex2009}, CIFAR-$10$ and CIFAR-$100$,
are subsets of the $80$ million tiny images~\cite{TorralbaFF08}.
Both datasets contain $60000$ $32\times32$ color images with $50000$ images for training and $10000$ images for test.
The CIFAR-$10$ dataset consists of $10$ classes, each of which contains $6000$ images.
There are $5000$ training images and $1000$ testing images per class.
The CIFAR-$100$ dataset consists of $ 100 $ classes, each of which contains $ 600 $ images.
There are $500$ training images and $100$ testing images per class.
The standard data augmentation scheme we adopt is widely used for
these datasets~\cite{HeZRS16, HuangSLSW16, LeeXGZT15, HuangLW16a, LarssonMS16a, LinCY13, RomeroBKCGB14, SpringenbergDBR14, SrivastavaGS15}:
we first
zero-pad the images with $4$ pixels on each side, and then
randomly crop them to produce $32\times32$ images, followed by
horizontally mirroring half of the images.
We normalize the
images by using the channel means
and standard deviations.

\begin{figure}
	\centering
	\includegraphics[width=0.45\textwidth]{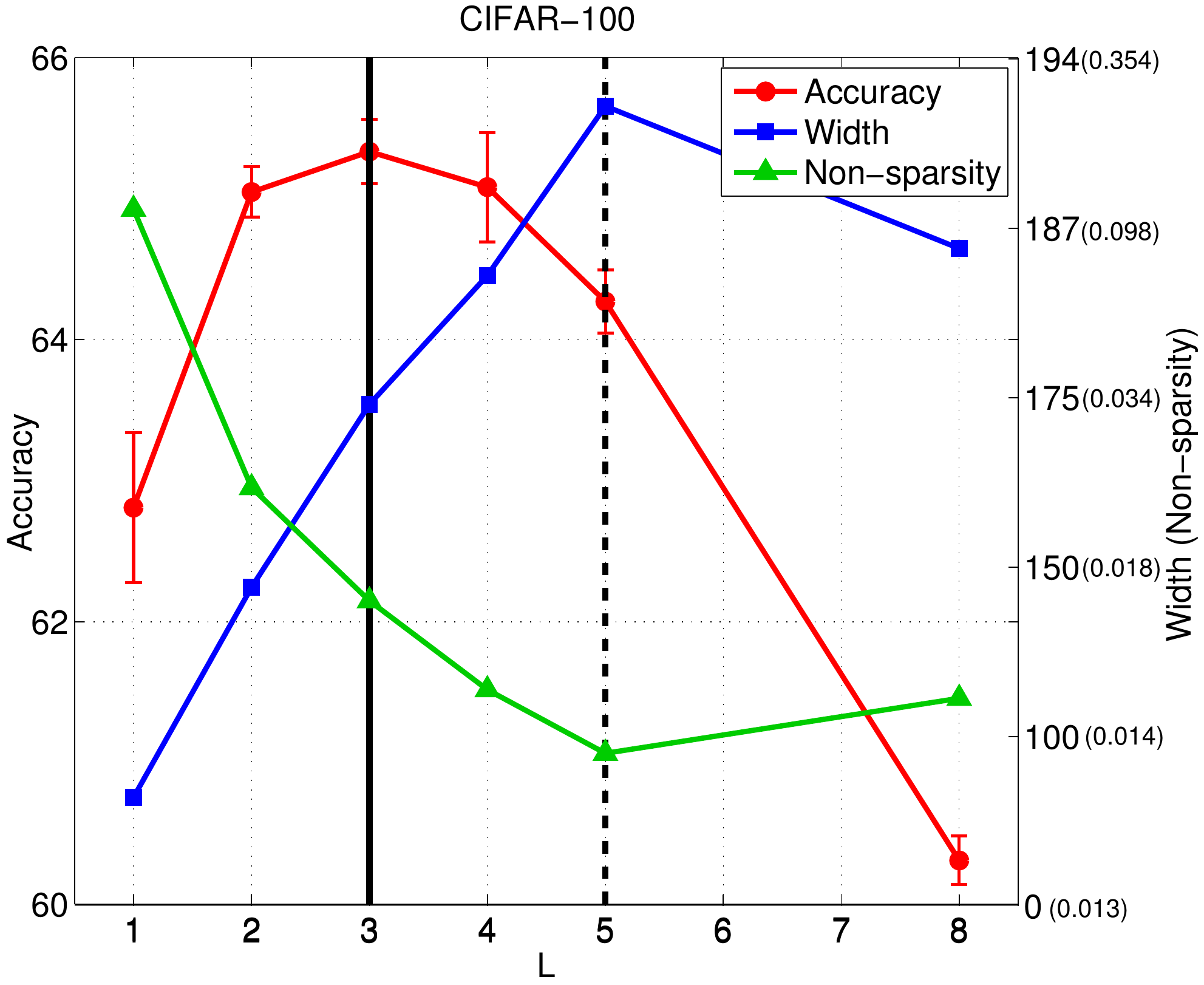}
	\caption{Illustrating how the number of layers $L$ affects the performance
		on CIFAR-$100$.
		The number of channels in each branch of group convolution
		is chosen to satisfy both the balance condition
		and the (nearly) complementary condition,
		and to keep the \#params the same.
		The maximum accuracy is achieved
		at some $L$,
		in which the width and the non-sparsity degree
		reach a balance.
	}
	\label{fig:theeffectofL}
	\vspace{-0.5cm}
\end{figure}

\begin{table*}
	\footnotesize
	\centering		
	\caption{Illustrating the architectures of networks we used in the experiments. {$x$ is the number of channels at the first stage. $B$ is the number of blocks and the skip connection is added every two blocks. $x\times(3\times3,1)$ means
		a $3\times 3$ channel-wise convolution with the channel number being $x$. $L$ and $K$ are the hyper-parameters of {IGCV$2$}. $[L-1,x,(1\times1, K)]$ denotes the ($L-1$) group $1\times 1$ convolutions with each branch containing $K$ channels.
	For IGCV$2$ ($Cx$), $L=3$, and for IGCV$2$* ($Cx$), $K=8$, $L^* = \lceil log_{K}(x) \rceil +1$.} }
	\label{tab:networkstructures}	
		\setlength\tabcolsep{5pt}
		\begin{tabular}{c|c|c|c|c}
			\hline
			Output size  & {Xception ($Cx$)} & {IGC-V1 ($Cx$)} & {IGCV$2$ ($Cx$)}   & {IGCV$2$* ($Cx$)} \\
			\hline			
			$32\times32$ &  ($3\times3$, $x$) & ($3\times3$, $ x $) &  ($3\times3$, $x$)  & ($3\times3$, $ 64 $) \\
			\hline
			$32\times32$ & $\begin{bmatrix}
			x\times(3\times3, 1) \\
			(1\times1, x)
			\end{bmatrix} \times B$  & $\begin{bmatrix}
			\frac x 2 \times(3\times3,2) \\
			2\times (1\times1, \frac x 2)
			\end{bmatrix} \times B$ & $\begin{bmatrix}
			x\times(3\times3,1)\\
			L-1, x, (1\times1,K_{s_1})
			\end{bmatrix} \times B$   & $\begin{bmatrix}
			x\times(3\times3, 1) \\
			L^*-1, x, (1\times1, K)
			\end{bmatrix} \times B$ \\
			\hline
			
			$16\times16$ &  $\begin{bmatrix}
			2x\times(3\times3,1) \\
			(1\times1,2x)
			\end{bmatrix} \times B$ & $\begin{bmatrix}
			x\times (3\times3 ,2) \\
			2\times (1\times1,x)
			\end{bmatrix} \times B$  & $\begin{bmatrix}
			2x\times( 3\times3,1)\\
			L-1, 2x,	(1\times1 , K_{s_2})
			\end{bmatrix} \times B$  & $\begin{bmatrix}
			2x\times(3\times3,1) \\
			L^*-1, 2x, (1\times1, K)
			\end{bmatrix} \times B$  \\
			\hline		
			
			$8~\times~8$ &  $\begin{bmatrix}
			4x\times(3\times3,1) \\
			(1\times1,4x)
			\end{bmatrix} \times B$ & $\begin{bmatrix}
			2x\times(3\times3,2)\\
			2\times (1\times1,2x)
			\end{bmatrix} \times B$ & $\begin{bmatrix}
			4x\times (3\times3,1) \\
			L-1, 4x, (1\times1, K_{s_3})
			\end{bmatrix} \times B$   & $\begin{bmatrix}
			4x\times (3\times3,1)\\
			L^*-1,4x,	(1\times1,K)
			\end{bmatrix} \times B$ \\
			\hline		
			
			$1\times1$ & \multicolumn{4}{c}{average pool, fc, softmax}  \\
			\hline			
			Depth	& \multicolumn{4}{c}{$3B+2$ }\\
			\hline				
		\end{tabular}
	\vspace{-0.3cm}
\end{table*}

\vspace{.1cm}
\noindent \textbf{Tiny ImageNet.}
The Tiny ImageNet dataset\footnote{https://tiny-imagenet.herokuapp.com/}
is a subset of ImageNet~\cite{ILSVRC15}.
The image size is resized
to $64 \times 64$.
There are $200$ classes, sampled
from $1000$ classes of ImageNet,
and $500$ training images, $50$ validation images
and $50$ testing images per class.
In our experiment,
we adopt the data augmentation scheme:
scale up the training images randomly to the size within $[64,80]$,
and randomly crop a $64\times 64$ patch for training, randomly horizontal mirroring
and normalize the cropped images
by subtracting the channel means and standard deviations.

\begin{figure}
	\centering
	\subfigure[]{\label{fig:deeperwider:a}\includegraphics[width=.48\linewidth, clip]{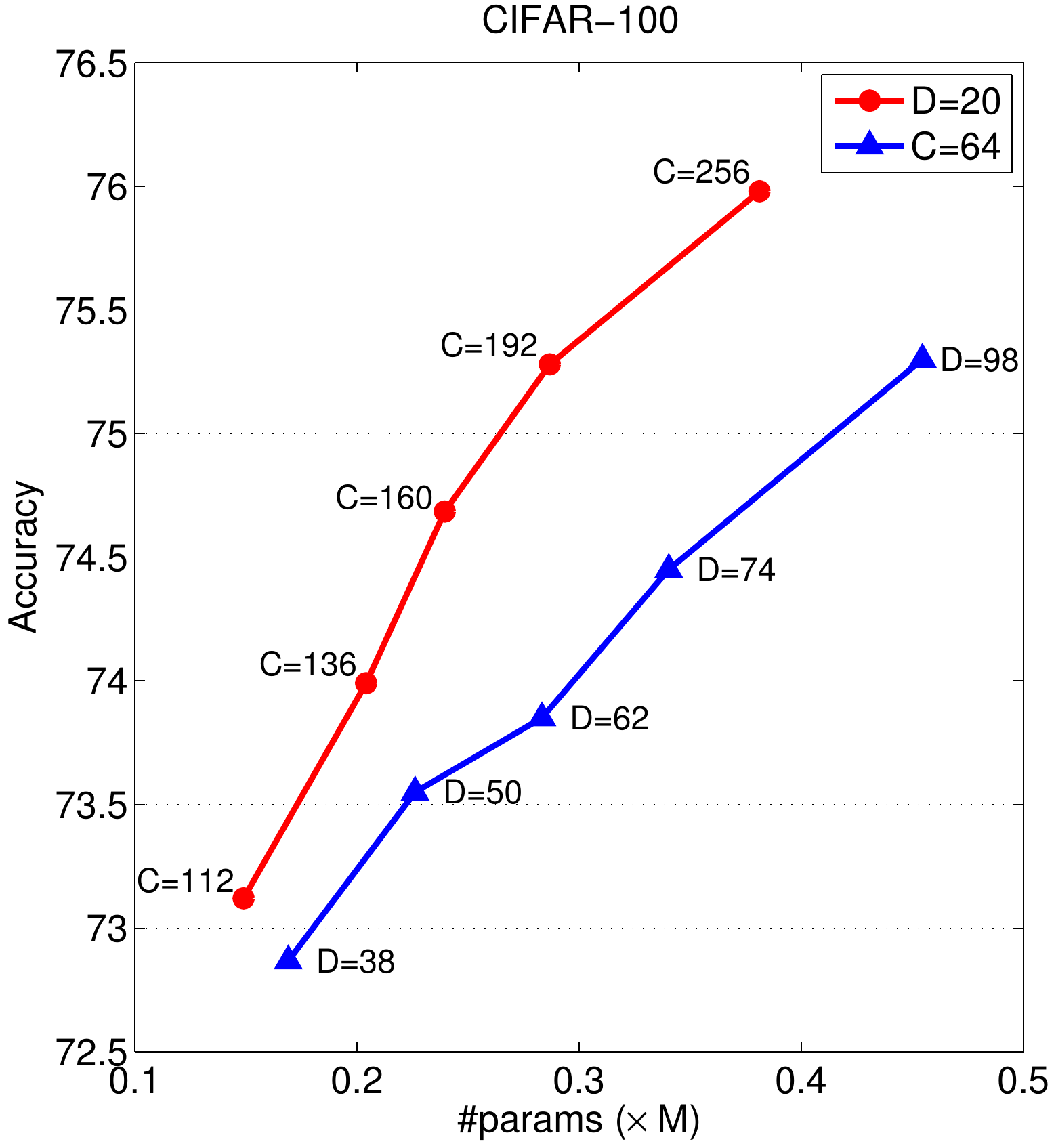}}~~
	\subfigure[]{\label{fig:deeperwider:b}\includegraphics[width=.48\linewidth, clip]{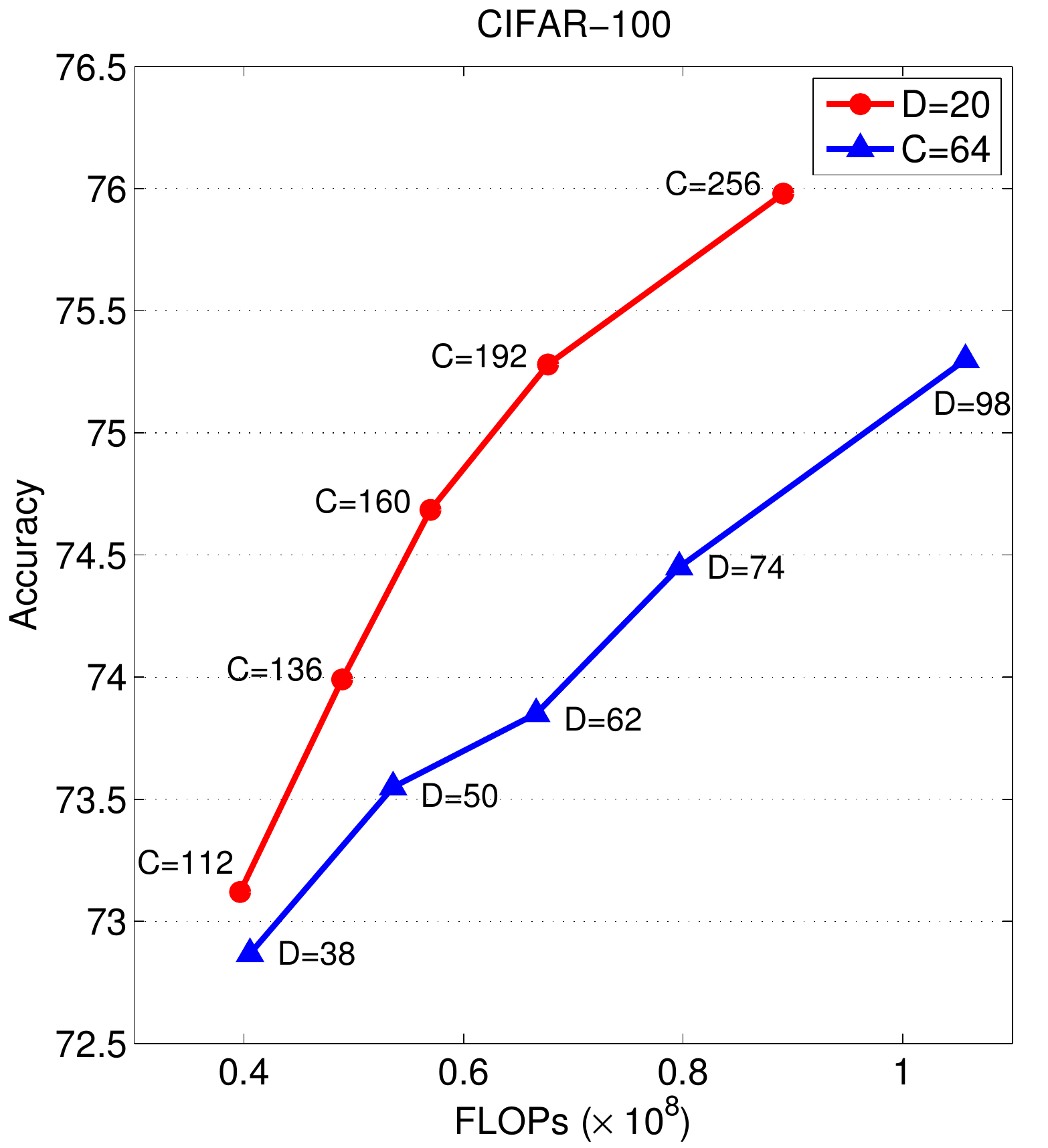}}
	
	\caption{Illustrating how the performance changes when our network goes deeper or wider.
		{We use the network structure IGCV2* ($Cx$)}
		in table \ref{tab:networkstructures} and conduct the experiments with various depths (denoted as $D$) and widths (denoted as $C$).
		Both going wider and going deeper increase the performance and the benefit of going wider is greater than going deeper.}
	\label{fig:deeperwider}
	\vspace{-0.5cm}
\end{figure}

\vspace{0.1cm}
\noindent \textbf{Training settings.}
For CIFAR, we adopt the same training settings as ~\cite{ZhangQXW17}.
We use SGD with Nesterov momentum to update network,
starting from learning rate 0.1 and
multiplying with a factor 0.1 at 200 epochs, 300 epochs and 350 epochs.
Weight decay is set as 0.0001 and momentum as 0.9.
We train the network with batch size as 64 for 400 epochs and report the accuracy at the final iteration.
The implementation is based on Caffe~\cite{jia2014caffe}.
For Tiny ImageNet, we use the similar training settings as CIFAR,
except that we train for totally 200 epochs and multiply the learning rate
with a factor 0.1 at 100 epochs, 150 epochs and 175 epochs.
To adapt Tiny ImageNet to the networks designed for CIFAR,
we set the stride of the first convolution layer as 2,
which is adopted in ~\cite{huang2017snapshot} as well.

\subsection{Empirical Analysis}

\vspace{.1cm}
\noindent\textbf{Complementary condition.}
We empirically investigate how the complementary condition
affects the performance
over the $8$-layer network without down-sampling,
where there are $6$ intermediate layers except the first convolutional layer
and the last FC layer.
We study an {IGCV$2$} building block,
which consists of $L$ group convolutions:
a channel-wise $3 \times 3$ convolution,
($L-1$) group $1 \times 1$ convolutions with each branch containing $K$ channels.
We conduct the studies over such blocks,
where the {IGCV$2$} is only applied to $1\times 1$ convolutions,
for removing the possible influence
of the coupling with spatial convolutions.
The results are given in Figure~\ref{fig:effectofk} with various values of
$K$ and $L$ under almost the same number of parameters.

An {IGCV$2$} block satisfying the complementary condition
leads to a dense kernel
with a large width
under the same number of parameters.
The {\color{red}red} bars in Figure~\ref{fig:effectofk}
depict the results for the {IGCV$2$} blocks that nearly satisfy this condition.
The {\color{blue}blue} bars in Figure~\ref{fig:effectofk} show the results
of the blocks
that correspond to dense kernels
but do not satisfy the complementary condition,
thus with more redundancy.
It can be seen that the networks with the {IGCV$2$} blocks \emph{(nearly)
	satisfying the complementary condition} (the red bars and the bars
next to the red bar in Figure~\ref{fig:effectofk} (a) and (c))
achieve the best performance.
We also notice that the red bar in Figure~\ref{fig:effectofk} (b),
satisfying the complementary condition
performs better than the best results
in Figure~\ref{fig:effectofk} (a) and (c).

In addition, we look at how the sparsity (density)
of the composed convolution kernel
affects the performance.
The results correspond to the {\color{green}green} bars in Figure~\ref{fig:effectofk}.
With $L$ being fixed, the kernel is more sparse with smaller $K$.
It can be seen that
the denser kernel leads to higher performance.
In Figure~\ref{fig:effectofk} (a),
the performance for the block $K=8$,
which is quite \emph{near to satisfy the complementary condition}
is slightly better than and almost the same to the denser kernel $K=12$.
The reason might be that
though the kernel for $K=8$ is more sparse,
but it corresponds to a larger width.

\vspace{.1cm}
\noindent\textbf{The effect of $L$.}
We empirically show how the number $L$ of group convolutions
affects the performance under the same number of parameters
on the CIFAR-$100$ dataset as an example.
We still use the $8$-layer network
and study the {IGCV$2$} block,
which contains a channel-wise spatial convolution
and $(L-1)$ group $1\times 1$ convolutions
satisfying the balance condition.

Figure~\ref{fig:theeffectofL}
shows the accuracy curve,
the width curve,
and the non-sparsity curve,
where the non-sparsity value
is the ratio of the number of non-zero parameters
to
the size of the resulting dense kernel matrix.
It can be seen that
the function of accuracy w.r.t. $L$
is concave,
the function of width w.r.t. $L$
is concave,
and the function of non-sparsity w.r.t. $L$
is convex.
The accuracy depends on both the width
and the non-sparsity degree.
When width becomes larger, accuracy might be higher.
On the other hand, accuracy might be lower when non-sparsity degree becomes smaller.
The black dashed line denotes an extreme case that the width is the largest and the non-sparsity is the smallest, the performance however is not the best. Instead,
the maximum accuracy is achieved at some $L$, in which the width and the non-sparsity degree reach a balance denoted by the black solid line.

\begin{table}
	\footnotesize
	\centering
	\caption{Classification accuracy comparison between Xception and {IGCV$2$} over the $ 8 $-layer network with various widths. The number of parameters and FLOPs are calculated within each block.}
	\label{tab:comparisontoxception8layers}	
	\setlength\tabcolsep{5pt}
	\begin{tabular}{l|c|c|cc}
		\hline
		\multirow{2}{*}{	Network} &\#Params / & \multirow{2}{*}{$C$} & \multirow{2}{*}{ CIFAR-$10$} & \multirow{2}{*}{CIFAR-$100$} \\
		& FLOPs & & & \\
		\hline
		Xception & $\approx 4700$ /	& $ 64 $  & $ 88.82\pm0.38$ & $62.81\pm0.53$ \\
		\cline{1-1} \cline{3-3}
		{IGCV$2$} & $\approx 4.8 \times 10^6$ & $ 144 $  & $\mathbf{89.12\pm0.15}$ & $\mathbf{65.05\pm0.18}$\\
		\hline
		Xception &$\approx 10000$ /& $ 96 $  & $ 90.21\pm0.62$ & $65.21\pm0.31$ \\
		\cline{1-1} \cline{3-3}
		{IGC-V$ 2 $} & $\approx 10^7$&$ 256 $  & $\mathbf{90.63\pm0.11}$ & $\mathbf{67.68\pm0.28}$\\
		\hline
		Xception &$\approx 17000$ /& $ 128 $ & $ 91.03\pm0.30$ & $66.51\pm0.68$ \\
		\cline{1-1} \cline{3-3}
		{IGC-V$ 2 $} & $\approx 1.7\times 10^7$ &$ 361 $ & $\mathbf{91.35\pm0.12}$ & $\mathbf{68.86\pm0.57}$\\			
		\hline
	\end{tabular}			
		\vspace{-0.1cm}
\end{table}
\vspace{.1cm}
\noindent\textbf{Deeper and wider networks.}
We also conduct experiments
to explore how the performance changes when our network
goes deeper and wider.
In this study,
the {IGCV$2$} block is composed
of a channel-wise $3\times 3$ convolution
and $(L-1)$ group $1\times 1$ convolutions.

We study the performance
over the networks with identity mappings
as skip connections,
where we replace the regular convolution in the residual network
with our {IGCV$2$} building block.
We do experiments on {IGCV$2$* ($Cx$) in Table~\ref{tab:networkstructures},
where $Cx$ means that the network width $C$ in the first stage is $x$}.
There are two types of experiments.
(1) Go wider: we fix the depth as 20 ($B=6$) and vary the width among
$\{112, 136, 160, 192, 256\}$.
(2) Go deeper:
we fix the width as $64$ and vary the depth among
$\{38, 50, 62, 74, 98\}$.
The results are shown in Figure~\ref{fig:deeperwider} (a) and (b).
One can see that
both going wider and going deeper
increase the network performance,
and the benefit of going wider using a relative small depth
is greater than that of going deeper,
which is consistent to the observation
for regular convolutions.

\begin{table}	
	\centering
	\footnotesize
	\caption{Classification accuracy comparison between Xception, IGC, and {our network}
		under various widths with depth fixed as $20$.}
	\label{tab:comparedIGCXceptionvariouswidth}	
		\setlength\tabcolsep{3.5pt}
		\begin{tabular}{l|cc|cc}
			\hline
			\multirow{2}{*}{Network}	& \#Params & FLOPs& \multirow{2}{*}{CIFAR-$ 100 $} &\multirow{2}{*}{ Tiny ImageNet} \\
			& $(\times M)$ & $(\times 10^8)$ & & \\
			\hline
			Xception ($C58$)  & $ 0.44 $ &$ 0.67 $ &$74.46\pm0.24$ &$57.34\pm0.34$ \\
			{IGCV$1$ ($C96$)}  & $ 0.40 $ &$ 0.82 $ & $75.10\pm0.36$ & $58.37\pm0.60$\\
			{IGCV$2$* ($C216$)}  & $ 0.32  $&$ 0.76 $ &$\mathbf{75.35\pm0.56}$ & $\mathbf{59.40\pm0.22}$\\
			\hline
			Xception ($C71$)  & $ 0.66 $ &$ 0.98 $ & $75.66\pm0.11$ &$58.28\pm0.16$ \\
			{IGCV$1$ ($C126$)}  &  $ 0.60 $ &$ 1.28 $ & $76.18\pm0.20$ &$59.67\pm0.47$\\
			{IGCV$2$* ($C304$)}  & $ 0.48 $ &$ 1.12 $  & $\mathbf{76.30\pm0.19}$ &$\mathbf{60.50\pm0.29}$\\			
			\hline
			Xception ($C83$)  & $ 0.89 $ &$ 1.31 $ &$76.22\pm0.05$& $59.01\pm0.52$ \\
			{IGCV$1$ ($C150$)}  & $  0.79 $ &$ 1.72 $ & $76.69\pm0.51$& $60.73\pm0.50$\\
			{IGCV$2$* ($C416$)}  & $ 0.65 $ &$ 1.52 $ &$\mathbf{77.02\pm0.20}$ &$\mathbf{60.98\pm0.23}$\\
			
			\hline
		\end{tabular}		
	\vspace{-0.5cm}
\end{table}

\subsection{Comparison With Xception}
\label{sec:comparisonwithxception}
We empirically show the comparison with Xception using various widths
and various depths under roughly the same number of parameters.

\vspace{.1cm}
\noindent\textbf{Varying the width.}
We firstly conduct the experiments
over the $8$-layer network without down-sampling,
where the $6$ intermediate convolutional layers
(except the first convolutional layer
and the last FC layer)
are replaced with Xception blocks and our {IGCV$2$} blocks.
Our {IGCV$2$} block is composed of a channel-wise $3 \times 3$ convolution similar to Xception, and
two group $1\times 1$ convolutions corresponding to the $1\times 1$ convolution in Xception.
The comparison on CIFAR-$10$ and CIFAR-$100$
is given in Table~\ref{tab:comparisontoxception8layers}.
It can be seen that
our networks consistently perform better than Xception
on both CIFAR-$10$ and CIFAR-$100$ datasets.
In particular on CIFAR-$100$, our networks achieve at least $2\%$ improvement.
The reason might be that our network using the {IGCV$2$} block is wider and thus improve the performance.

In addition,
we report the results over
$20$-layer networks with various widths.
The networks we used are
{IGCV$2$* ($Cx$) and Xception ($Cx$)} illustrated in Table~\ref{tab:networkstructures},
and we fix the channel number at the first convolutional layer in Xception as $35$.
The results are presented in Table~\ref{tab:comparedIGCXceptionvariouswidth}.
We can see that our network with fewer number of parameters, performs better than Xception,
which shows the powerfulness of {IGCV$2$} block.

\vspace{.1cm}
\noindent\textbf{Varying the depth.}
We also compare the performances
between Xception and our network {IGCV$2$}
with various depths:
$8, 20, 26$.
The width $C$ in Xception is fixed as $35$,
and the width in our network is set to $64$ in order to keep the
number of parameters smaller than Xception.
For {IGCV$2$ ($C64$)}, the number of channels in each branch are the same within the stage and
are different for the three stages. To
satisfy the complementary condition, $K_{s_1}$, $K_{s_2}$, and $K_{s_3}$ are set to
$8, 16, 32$ respectively.
The results over
CIFAR-$100$ and Tiny ImageNet
are given in Table~\ref{tab:comparisontoXceptionvariousdepth}.
Our {IGCV$2$ ($C64$) network
performs better than Xception ($C35$)}
with smaller numbers of parameters
and smaller or similar computation complexity.
For example, when depth is $26$, our network consuming fewer number of parameters and less computation complexity gets $56.32\%$ accuracy on Tiny ImageNet, $1\%$ better than $55.39\%$ of Xception.

\begin{table}
	\centering
	\caption{Classification accuracy comparison between Xception and our network under various depths.}
	\label{tab:comparisontoXceptionvariousdepth}

		\scriptsize
		\setlength\tabcolsep{5pt}
		\begin{tabular}{c|cc|cc}
			\hline
			\multirow{2}{*}{ Network}	& \#Params & FLOPs & \multirow{2}{*}{CIFAR-$100$} & \multirow{2}{*}{Tiny ImageNet} \\			
			& $(\times M)$ & $(\times 10^8)$ & & \\
			\hline
			\multicolumn{5}{c}{D$=8$}\\
			\hline
			Xception ($C35$) & $ 0.056 $ & $ 0.095 $ & $67.00\pm0.29$ & $49.91\pm0.18$\\
			{IGCV$2$ ($C64$)} & $ 0.047 $ & $ 0.095 $ & $\mathbf{67.83\pm0.35}$ & $\mathbf{51.40\pm0.32}$\\
			\hline
			\multicolumn{5}{c}{D$=20$}\\
			\hline
			Xception ($C35$) & $ 0.168 $ & $ 0.268 $ & $70.97\pm0.10$ & $55.29\pm0.19$\\
			{IGCV$2$ ($C64$)} & $ 0.149 $ & $ 0.262 $ &$\mathbf{71.69\pm0.09}$ & $\mathbf{55.72\pm0.43}$\\
			\hline
			\multicolumn{5}{c}{D$=26$}\\
			\hline
			Xception ($C35$) & $ 0.223 $ & $ 0.355 $ & $72.48\pm0.18$ &  $55.39\pm0.35$ \\
			{IGCV$2$ ($C64$)} & $ 0.200 $ & $ 0.346 $ & $\mathbf{72.94\pm0.26}$ & $\mathbf{56.32\pm0.38}$\\
			\hline   		    			
		\end{tabular}	
	\vspace{-0.5cm}
\end{table}

\subsection{Comparison With IGC.}
\label{sec:comparisonwithIGC}
\noindent\textbf{Varying the width.}
Similar to the comparison with Xception,
we perform comparison with IGC~\cite{ZhangQXW17}
(denoted by IGCV$1$ in experiments)
over the simple networks
and over $20$-layer networks with different widths.
The results are presented respectively in
Table~\ref{tab:comparisontoIGC8layers} and
Table~\ref{tab:comparedIGCXceptionvariouswidth},
in which the observation
is consistent to that
from the comparison to Xception.

Let us look at the detailed results
over the simple $8$-layer network.
The {IGCV$1$} block is designed by following~\cite{ZhangQXW17}:
the primary group convolution contains two branches.
We study two designs of our {IGCV$2$} block.
The first design follows the {IGCV$1$} design manner:
the first group convolution contains two branches
and the other two group convolution contains
the same number of branches.
In this case, $SK_1 = 18$,
$K_2 = K_3 = 7$ ($9$, $11$ for other two bigger networks),
which is far from the balance condition
given in Equation~\ref{eqn:balancecondition}.
In the second design,
we use a channel-wise $3 \times 3$ convolution,
two group $1\times 1$ convolutions.
In this case,
$SK_1 = 9$,
$K_2 = K_3 = 10$ ($13$, $16$ for other two bigger networks),
which satisfies the balance condition.
The resulting networks are denoted as {IGCV$2$ I} and {IGCV$2$ II}
respectively.
The results given in Table~\ref{tab:comparisontoIGC8layers} show (i) that the first design performs better than {IGCV$1$} on CIFAR-$100$
and a little worse than {IGCV$1$} on CIFAR-$10$,
which might stem from
different balance condition satisfaction degrees,
and (ii) that the second design performs the best,
which stems from the better satisfaction
with the balance condition.

In addition,
the comparison over $20$-layer network shown in Table~\ref{tab:comparedIGCXceptionvariouswidth}
verifies that our network {IGCV$2$} with
smaller model size as well as less computation complexity, is able to
achieve better performance under various widths.

\begin{table}
	\scriptsize
	\centering
	\caption{Classification accuracy comparison between {IGCV$1$} and our networks with two designs, {IGCV$2$ I} and {IGCV$2$ II}, over the $8$-layer network with various widths.
		The number of parameters and FLOPs are calculated within each block.}
	\label{tab:comparisontoIGC8layers}	
	\setlength\tabcolsep{5.5pt} 	
	\begin{tabular}{l|c|c|cc}
		\hline
		\multirow{2}{*}{Network}& \#Params / & \multirow{2}{*}{$C$} & \multirow{2}{*}{ CIFAR-$ 10 $} & \multirow{2}{*}{ CIFAR-$ 100 $} \\
		& FLOPs & & & \\
		\hline
		{IGCV$1$} & \multirow{3}{*}{$\begin{array}{c}
			\approx 3000~ /\\
			\approx 3.2\times 10^6\\
			\end{array}$} &$ 64 $ &$87.93\pm0.10$ &$60.89\pm0.46$ \\
		{IGCV$2$ I} & & $ 98 $  & $87.41\pm0.35$ & $61.35\pm0.41$\\
		{IGCV$2$ II} & & $ 100 $ &$\mathbf{88.01\pm 0.32}$ & $\mathbf{62.32\pm0.57}$\\
		\hline
		{IGCV$1$} &\multirow{3}{*}{$\begin{array}{c}
			\approx 6000~ /\\
			\approx 6.1\times 10^6\\
			\end{array}$} & $ 96 $  & $89.29\pm0.26$ &$64.58\pm0.19$ \\
		{IGCV$2$ I} & & $ 162 $  & $89.06\pm0.08$ &$65.39\pm0.61$\\
		{IGCV$2$ II} && $ 169 $  &$\mathbf{89.66\pm0.17}$ &$\mathbf{66.00\pm0.78}$\\			
		\hline
		{IGCV$1$} & \multirow{3}{*}{$\begin{array}{c}
			\approx 10000~ /\\
			\approx 1.1\times 10^7\\
			\end{array}$}& $ 128 $  &$90.16\pm0.20$ & $66.59\pm0.48$ \\
		{IGCV$2$ I} && $ 242 $  & $89.85\pm0.22$ & $67.21\pm0.34$\\
		{IGCV$2$ II} & &$ 256 $  & $\mathbf{90.63\pm0.11}$ &$\mathbf{67.68\pm0.28}$\\
		
		\hline
	\end{tabular}				
	\vspace{-0.2cm}
\end{table}

\begin{table}
	\centering
	\caption{Classification accuracy comparison between {IGCV$1$} and our network under various depths.}
	\label{tab:comparisontoIGCvariousdepth}		
	
	\scriptsize
			\setlength\tabcolsep{4pt} 	
	\begin{tabular}{c|cc|cc}
		\hline
		\multirow{2}{*}{ Network}	& \#Params & FLOPs & \multirow{2}{*}{CIFAR-$100$} & \multirow{2}{*}{Tiny ImageNet} \\			
		& $(\times M)$ & $(\times 10^7)$ & & \\
		\hline
		\multicolumn{5}{c}{D$=8$}\\
		\hline
		{IGCV$1$ ($C48$)} & $ 0.046 $ & $ 1.00 $ & $66.52\pm0.12$ & $48.57\pm0.53$\\
		{IGCV$2$ ($C80$)} & $ 0.046 $ & $ 1.18 $ & $\mathbf{67.65\pm0.29}$ & $\mathbf{51.49\pm0.33}$\\
		\hline
		\multicolumn{5}{c}{D$=20$}\\
		\hline
		{IGCV$1$ ($C48$)} & $ 0.151 $ & $ 2.89 $ & $70.87\pm0.39$ & $54.83\pm0.27$\\
		{IGCV$2$ ($C80$)} & $ 0.144 $ & $ 3.20 $ &$\mathbf{72.63\pm0.07}$ & $\mathbf{56.40\pm0.17}$\\
		\hline
		\multicolumn{5}{c}{D$=26$}\\
		\hline
		{IGCV$1$ ($C48$)} & $ 0.203 $ & $ 3.83 $ & $71.82\pm0.40$ &  $56.64\pm0.15$ \\
		{IGCV$2$ ($C80$)} & $ 0.193 $ & $ 4.21 $ & $\mathbf{73.49\pm0.34}$ & $\mathbf{57.12\pm0.09}$\\
		\hline   		    			
	\end{tabular}	
	\vspace{-0.5cm}
\end{table}
\vspace{.1cm}
\noindent\textbf{Varying the depth.}
We also show the performance comparison between {IGCV$1$} and our network
under various depths: $8, 20, 26$.
The width $C$ is set to $48$ in {IGCV$1$} (corresponding to IGC-$L24M2$ in~\cite{ZhangQXW17}) and the width in our network is set to $80$.
The network structure can be seen in Table~\ref{tab:networkstructures} and
here $K_{s_1}, K_{s_2}, K_{s_3}$ are set to $10,16,20$ respectively to satisfy the complementary condition.
The results over CIFAR-$100$ and Tiny ImageNet
are given in Table~\ref{tab:comparisontoIGCvariousdepth},
showing superior performance of our network over {IGCV$1$}, which
demonstrates the effectiveness of {IGCV$2$} block.

\subsection{Performance Comparison to Small Models}
We show the advantages of our network
with small models
by comparing to
existing state-of-the-art architecture
design algorithms.
The results are presented in Table~\ref{tab:smallmodelcomparison}.
The observation is that our network with a smaller model
achieves similar classification accuracies.
For example, on CIFAR-$100$, the classification error of our approach
with $0.65M$ parameters
is $22.95\%$,
while
Swapout~\cite{SinghHF16} reaches $22.72\%$
with $7.4M$ parameters,
FractalNet~\cite{LarssonMS16a} reaches $23.30\%$
with $38.6M$ parameters, WRN-$40$-$4$~\cite{ZagoruykoK16} reaches $21.18$ with $8.9M$ parameters and
WRN-$32$-$4$~\cite{huang2017snapshot} reaches $23.55\%$
with $7.4M$ parameters.
On Tiny ImageNet, our network contains
the smallest number of parameters,
and achieves better performance
compared to the reported results.
On CIFAR-$10$,
our network achieves competitive performance.
In table \ref{tab:smallmodelcomparison}, DenseNet-BC($k=12$)~\cite{HuangLW16a} with more number of parameters achieves lower classification error on CIFAR-$100$ and CIFAR-$10$ compared with {IGCV$2$* (C$416$)}. Dense connection is a structure complementary to {IGCV$1$}, and we can also combine densely connected structure with {IGCV$1$} to improve the performance.  We believe that
our approach potentially gets more improvement
if dense connection and bottleneck design are exploited.

\begin{table}
	\centering
	\scriptsize
	\caption{Illustrating
		the advantages of our networks for the small model cases
		through the classification error comparison to existing state-of-the-art architecture design algorithms.}
	\label{tab:smallmodelcomparison}
	\setlength\tabcolsep{2.5pt} 		
	\begin{tabular}{l|c|c|c|c|c}			
		
		\hline
		& D & \#Params($M$) & C$10$ & C$100$ & Tiny ImageNet \\
		\hline
		Swapout~\cite{SinghHF16} & $20$ & $1.1$ & $6.58$ & $25.86$ & -\\
		& $32$ & $7.4$ & $4.76$&$22.72$ &- \\
		\hline
		DFN~\cite{WangWZZ16} & $50$ &$3.7$ &$6.40$ & $27.61$ &- \\
		& $50$ & $3.9$ & $6.24$ &$27.52$ &- \\
		\hline
		FractalNet~\cite{LarssonMS16a}& $21$&$38.6$ &$5.22$ & $23.30$ &- \\
		
		\hline
		ResNet~\cite{HuangSLSW16} & $110$ & $1.7$ & $6.41$ & $27.76$ & -\\
		
		ResNet(pre-act) ~\cite{he2016identity} & $164$ & $1.7$ & $5.46$ & $24.33$ &- \\
		ResNet~\cite{huang2017snapshot}&$110$&$1.7$& $5.52$ &$28.02$&$46.5$\\			
		
		\hline
		DFN-MR1~\cite{ZhaoWLTZ16} & $56$ & $1.7$ & $4.94$ & $24.46$ &-\\
		\hline
		RiR~\cite{TargAL16}& $18$ & $10.3$ & $5.01$ & $22.90$ & - \\
		\hline
		ResNet$34$~\cite{cordeirowide} & $34$ & $21.4$ & &- &$46.9$ \\
		ResNet$18$-$2\times$~\cite{cordeirowide}& $18$ & $25.7$ & & - &$44.6$ \\
		\hline
		WRN-$32$-$4$~\cite{huang2017snapshot} & $32$ & $7.4$ & $5.43$ & $23.55$ & $39.63$\\	
		WRN-$40$-$4$~\cite{ZagoruykoK16} & $40$ & $8.9$ & $4.53$  & $21.18$ & - \\		
		\hline
		DenseNet($k=12$)~\cite{huang2017snapshot} & $40$ & $1.0$ &- & - & $39.09$ \\
		DenseNet($k=12$)~\cite{HuangLW16a} &  $40$ & $1.0$ & $5.24$ & $24.42$ &- \\
		DenseNet-BC($k=12$)~\cite{HuangLW16a}& $100$ & $0.8$ & $4.51$ & $22.27$ &- \\
		\hline			
		{IGCV$2$}*-$C416$ & $20$ & $0.65$ & $5.49$& $22.95$ & $38.81$ \\			
		\hline
		
	\end{tabular}
\end{table}

\section{Comparison to MobileNet on ImageNet}

We compare our approach to MobileNetV$1$~\cite{howard2017mobilenets}
and MobileNetV$2$~\cite{SHZZC18}
on the ImageNet classification task~\cite{ILSVRC15}.
We use SGD to train the networks
using the same hyperparameters
(weight decay $=0.00004$, and momentum $=0.9$).
The mini-batch size is $96$,
and we use $4$ GPUs ($24$ samples per GPU).
We adopt the same data augmentation as in~\cite{howard2017mobilenets, SHZZC18}.
We train the models for $100$ epochs with extra $20$ epochs for retraining on MXNet~\cite{ChenLLLWWXXZZ15}.
We start from a learning rate of $0.045$,
and then divide it by $1$ every $30$ epochs.
We evaluate on the single $224\times224$ center crop
from an image whose shorter side is $256$.

\subsection{IGCV$2$ vs. MobileNetV$1$}
\begin{figure}
	\centering
	\includegraphics[width=0.4\textwidth]{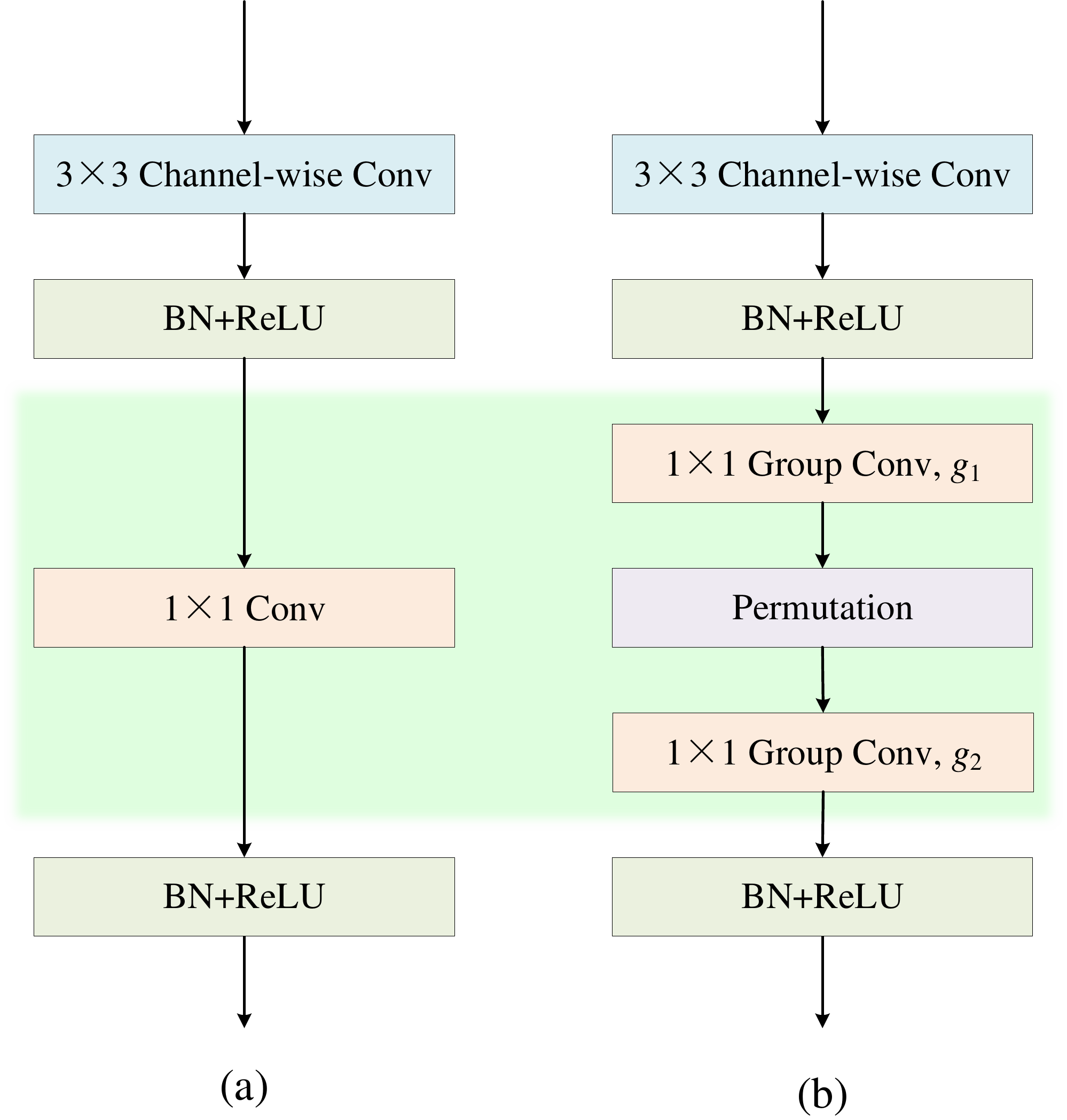}
	\caption{(a) A block in MobileNetV$1$.
(b) A nonlinear IGCV$2$ block.}
	\label{fig:MobileNetv1blocktoIGCV2}
\end{figure}

\begin{table*}
	\centering
\footnotesize
	\caption{The IGCV$2$ network and
MobileNetV$1$.
$c_i$ and $c_o$ are the number of input and output channels
for the blocks in the corresponding lines.
$g_1$ and $g_2$
are the number of branches in the two $1\times 1$ group convolutions.
Blocks with $*$ means that there is no $3\times3$ channel-wise convolution.}
	\label{tab:IGCV21.0}
	\begin{tabular}{c|c|l|c|c|c|c|l|c|c}
		\hline
		& &\multicolumn{5}{c|}{IGCV2-$1.0$} & \multicolumn{3}{c}{MobileNetV1-$1.0$}\\
		\hline
		Input size& Stride &Block & $c_i$ & $c_o$ & $g_1$ & $g_2$ & Block & $c_i$ & $c_o$ \\
		\hline		
		$224\times 224$& $2$ &Conv$2d$ ~$\times 1$ & $3$ & $64$ & - & - & Conv$2d$ ~$\times 1$ & $3$ & $32$ \\
		\hline		
		$112\times 112$& $1$ &IGCV2 ~$\times 1$ & $64$ & $128$ & $8$ & $8$  & Xception ~$\times 1$ & $32$ & $64$ \\
		\hline		
		$112\times 112$& $2$&IGCV2 ~$\times 1$ & $128$& $256$ & $8$ & $8$ &  Xception ~$\times 1$& $64$ & $128$\\
		\hline		
		$56\times 56$& $1$&IGCV2 ~$\times 1$ & $256$& $256$ & $8$ & $8$ &  Xception ~$\times 1$& $128$ & $128$\\
		\hline		
        $56\times 56$& $2$&IGCV2 ~$\times 1$ & $256$& $512$ & $8$ & $8$ &  Xception ~$\times 1$& $128$ & $256$\\
		\hline
        $28\times 28$& $1$&IGCV2 ~$\times 1$ & $512$ & $512$ &$8$ & $8$ &  Xception ~$\times 1$& $256$& $256$\\
		\hline
        $28\times 28$& $2$&IGCV2 ~$\times 1$ & $512$ & $1024$& $8$& $8$ &  Xception ~$\times 1$& $256$& $512$\\
		\hline
        $14\times 14$& $1$&IGCV2 ~$\times 10$ & $1024$& $1024$ &$8$ & $8$&  Xception ~$\times 10$& $512$& $512$\\	
		\hline
		$14\times 14$& $2$&IGCV2 ~$\times 1$ &$1024$ & $2048$ & $8$&$8$ &  Xception ~$\times 1$& $512$&$1024$\\
		\hline		
        $7\times 7$& $1$&IGCV2 ~$\times 1$ & $2048$ & $2048$ &$8$ &$8$ &  Xception ~$\times 1$& $1024$&$1024$ \\		
		\hline
		$7\times 7$ & $1$& \multicolumn{5}{c}{Global average pooling} & \multicolumn{3}{c}{Global average pooling} \\
		 \hline
        $1\times 1$& $1$ &IGCV2* ~$\times 1$ & $2048$ & $1000$ & $8$ & $4$ & FC & $1024$ &$1000$\\		
		\hline
 	\end{tabular}
\end{table*}

\begin{table}[t]
	\footnotesize
	\centering		
	\caption{A comparison of MobileNetV$1$ and IGCV$2$ on ImageNet classification. ${1.0,0.5,0.25}$ are width multipliers.}
	\label{tab:compareIGCV2obilenetv1}	
	\begin{tabular}{l|c|c|c}
		\hline
		Network & \#Params (M) & FLOPs (M) & Accuracy ($\%$)\\
		\hline
		MobileNet-$1.0$ & $4.2$ & $569$ & $70.6$ \\
		IGCV2-$1.0$ & $4.1$ & $564$ & $\bf 70.7$ \\ 	
		\hline
		MobileNet-$0.5$ & $1.3$ & $149$ & $63.7$ \\
		IGCV2-$0.5$ & $1.3$ & $156$ & $\bf65.5$ \\
		\hline
		MobileNet-$0.25$ & $0.5$ & $41$ & $50.6$ \\
		IGCV2-$0.25$ & $0.5$ & $46$ & $\bf54.9$ \\ 		
		\hline		
	\end{tabular}
\end{table}

We form our network using the same pattern as MobileNetV$1$~\cite{howard2017mobilenets}:
same number of blocks, no skip connections.
In particular, we use a nonlinear IGCV$2$:
$3\times 3$ channel-wise convolution $\rightarrow$
$\operatorname{BN}\rightarrow \operatorname{ReLU} \rightarrow$
$1\times 1$ group convolution $\rightarrow$
$\operatorname{BN}\rightarrow$
$\operatorname{Permutation}$ $\rightarrow$
$1\times 1$ group convolution $\rightarrow$
$\operatorname{BN}\rightarrow \operatorname{ReLU} \rightarrow$
$\operatorname{Permutation}$,
which is illustrated in Figure~\ref{fig:MobileNetv1blocktoIGCV2}.
The dimension increase, if included, is conducted
over the last $1\times 1$ group convolution.
We adopt a loose complementary condition to form an IGCV$2$ block:
each $1\times 1$ group convolution contains $8$ ($g_1 = g_2 = 8$) branches,
in which some channels in the same branch might still lie in the same branch
in another group convolution.
The description of the IGCV$2$-$1.0$ and MobileNet-$1.0$ networks with the same number of parameters are
shown in Table~\ref{tab:IGCV21.0}.
The results are given in Table \ref{tab:compareIGCV2obilenetv1}.
This result demonstrates that IGCV2 is effective as well on large scale image dataset.

\subsection{IGCV$3$ vs. MobileNetV$2$}
\begin{figure}
	\centering
	\includegraphics[width=0.4\textwidth]{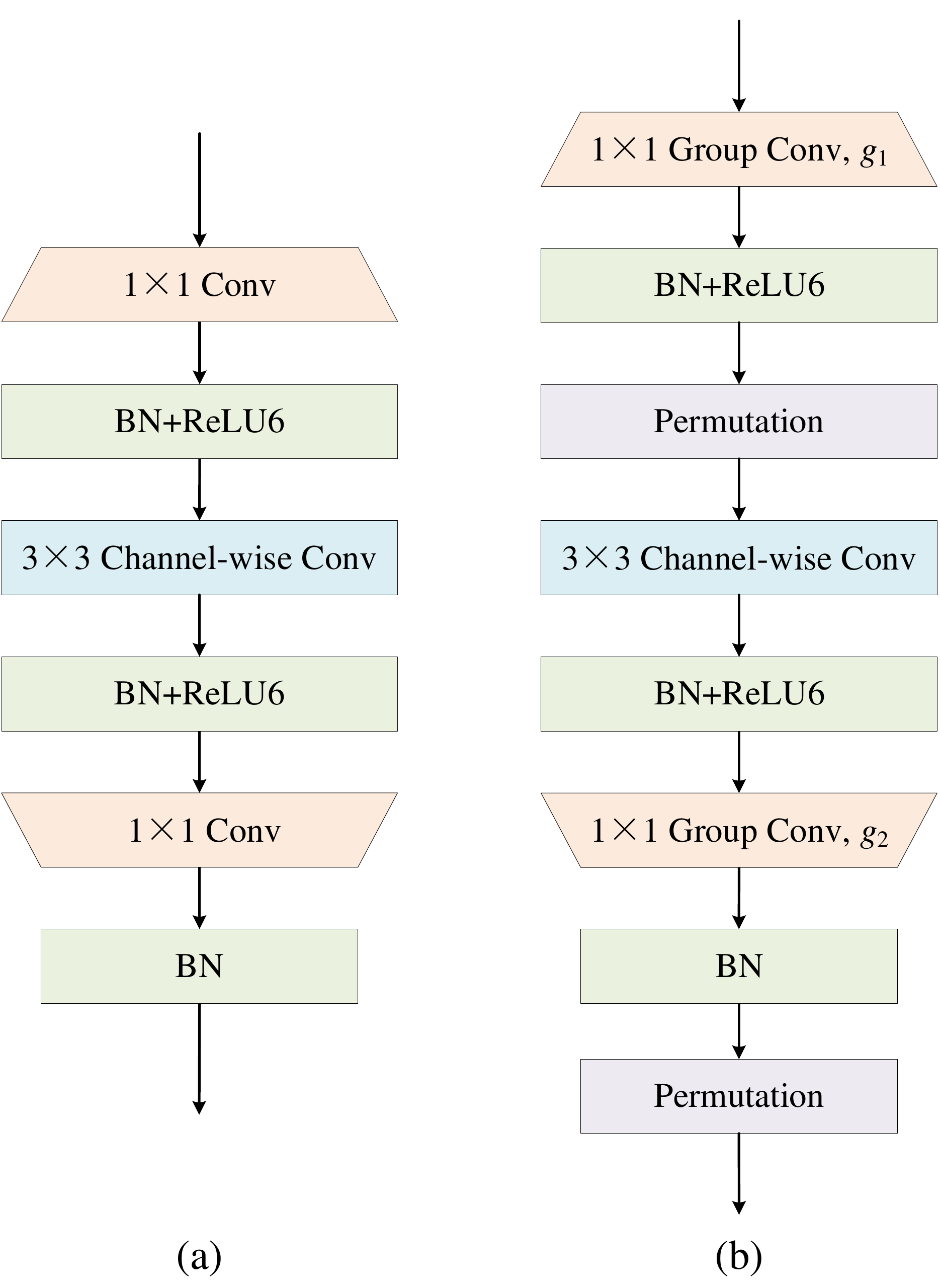}
	\caption{(a) A block in MobileNetV$2$.
(b) A nonlinear IGCV$3$ block.}
	\label{fig:IGCV3}
\end{figure}

We introduce an IGCV$3$ block:
Combine the low-rank convolution kernels, bottleneck,
and IGCV$2$,
which is illustrated in Figure~\ref{fig:IGCV3}.
We adopt nonlinear IGCV$3$ blocks
and form it with the loose complementary condition:
each $1\times 1$ group convolution contains $2$ ($g_1 = g_2 = 8$) branches.
In the constructed networks,
there is a skip connection for each block except  the downsampling blocks,
and two IGCV$3$ blocks correspond to one block
MobileNetV$2$~\cite{SHZZC18}.
The comparison results are given in Table \ref{tab:cmp-mobilenetv2}.

\begin{table*}
	\centering
\footnotesize
	\caption{The IGCV$3$ network and
MobileNetV$2$. $t$ is the expansion factor. $c_i$ and $c_o$ are the number of input and output channels
for the blocks in the corresponding lines.
$g_1$ and $g_2$
are the number of branches in the two $1\times 1$ group convolutions.
Blocks with $*$ means that there is no $3\times3$ channel-wise convolution.}.
	\label{tab:IGCV3}
	\begin{tabular}{c|c|c|l|c|c|c|c|l|c|c}
		\hline
		& & &\multicolumn{5}{c|}{IGCV$3$-$1.0$} & \multicolumn{3}{c}{MobileNetV$2$-$1.0$}\\
		\hline
		Input size& Stride& $t$ &Block & $c_i$ & $c_o$ &$g_1$&$g_2$& Block & $c_i$ & $c_o$ \\
		\hline		
		$224^2$& $2$ &$-$ &Conv$2d$ ~$\times 1$ & $3$ & $32$&$-$&$-$ & Conv$2d$ ~$\times 1$ & $3$ & $32$ \\
		\hline		
		$112^2$& $1$ & $1$ &IGCV$3$ ~$\times 1$ & $32$ & $16$&$2$&$2$& Bottleneck ~$\times 1$ & $32$ & $16$ \\
		\hline	
		$112^2$& $2$ & $6$&IGCV$3$ ~$\times 1$ & $16$& $24$&$2$&$2$  &  Bottleneck ~$\times 1$& $16$ & $24$\\
		\hline
		$56^2$& $1$ & $6$&IGCV$3$ ~$\times 3$ & $24$& $24$&$2$&$2$ &  Bottleneck ~$\times 1$& $24$ & $24$\\
		\hline
		$56^2$& $2$ & $6$&IGCV$3$ ~$\times 1$ & $24$& $32$&$2$&$2$ &  Bottleneck ~$\times 1$& $24$ & $32$\\
		\hline
		$28^2$& $1$ & $6$&IGCV$3$ ~$\times 5$ & $32$& $32$&$2$&$2$ &  Bottleneck ~$\times 2$& $32$ & $32$\\
		\hline
        $28^2$& $2$ &$6$ &IGCV$3$ ~$\times 1$ &$32$ &$64$&$2$&$2$&  Bottleneck ~$\times 1$& $32$&  $64$\\
		\hline
		$14^2$& $1$ &$6$&IGCV$3$ ~$\times 7$ & $64$&  $64$&$2$&$2$  &  Bottleneck ~$\times 3$& $64$&  $64$\\
		\hline
        $14^2$& $1$ & $6$&IGCV$3$ ~$\times 1$ & $64$& $96$&$2$&$2$&  Bottleneck ~$\times 1$&$64$& $96$\\	
		\hline
		$14^2$& $1$ & $6$&IGCV$3$ ~$\times 5$ &$96$& $96$&$2$&$2$&  Bottleneck ~$\times 2$&$96$& $96$\\	
		\hline
		$14^2$& $2$ &$6$&IGCV$3$ ~$\times 1$ &$96$&$160$&$2$&$2$ &  Bottleneck ~$\times 1$& $96$&$160$\\
		\hline
		$7^2$& $1$ &$6$&IGCV$3$ ~$\times 5$ & $160$&$160$&$2$&$2$ &  Bottleneck ~$\times 2$& $160$&$160$\\
		\hline
        $7^2$& $1$ &$6$&IGCV$3$ ~$\times 1$ &  $160$&$320$&$2$&$2$&  Bottleneck ~$\times 1$& $160$&$320$ \\		
		\hline
		$7^2$& $1$ &$-$&IGCV$3$* ~$\times 1$ &  $320$&$1280$&$2$&$2$   &  Conv$2d$ ~$\times 1$& $320$&$1280$ \\		
		\hline
		$7^2$ & $1$&$-$ & \multicolumn{5}{c}{Global average pooling}& \multicolumn{3}{c}{Global average pooling} \\
		 \hline
        $1^2$& $1$ &$-$ &Conv$2d\times 1$& $1280$ & $1000$&$-$&$-$& Conv$2d\times 1$& $1280$ &$1000$\\		
		\hline
 	\end{tabular}
\end{table*}

\begin{table}
	\footnotesize
	\centering
	\setlength{\tabcolsep}{1mm}
	\caption{A comparison of MobileNetV$2$ and IGCV$3$ on ImageNet classification. ${0.7, 1.0}$ are width multipliers.}
	\label{tab:cmp-mobilenetv2}	
	\begin{tabular}{l|c|c|c}
		\hline	
		Network & \#Params (M) & FLOPs (M) & Accuracy ($\%$)\\
		\hline
		MobileNetV2-0.7 (Our impl.) & $1.9$ & $160$ &$66.57$\\
		IGCV$3$-0.7 ({ours}) & $2.0$ & $170$ &$\bf 68.46$\\
		\hline
MobileNetV1 & $4.2$ & $569$ & $70.6$ \\
		MobileNetV2-1.0~\cite{SHZZC18} & $3.4$ & $300$ &$72.0$\\
		MobileNetV2-1.0 (Our impl.) & $3.4$ & $300$ &$71.0$\\
		IGCV$3$ ({ours}) & $3.5$ & $320$ &$\bf 72.2$\\
		\hline
	\end{tabular}
\end{table}
\section{Conclusion}
In this paper,
we aim to eliminate the redundancy
in convolution kernels
and present an Interleaved Structured Sparse Convolution (IGCV$2$) block
to compose a dense kernel.
We present the complementary condition and the balance condition to guide
the design
and obtain a balance among model size, computation complexity and classification accuracy.
Empirical results show the advantage over MobileNetV1 and MobileNetV2,
and demonstrate that our network with smaller model size achieves similar performance compared with other network structures.

\section*{Acknowledgement}
We appreciate Ke Sun, Mingjie Li, and Depu Meng for helping
on the experiments
about the comparison to MobileNetV$1$
and MobileNetV$2$.
{\small
	\bibliographystyle{ieee}
	\bibliography{ConvApproximation}
}

\end{document}